\documentclass{article}

\usepackage[final]{neurips_data_2021}

\usepackage[utf8]{inputenc} 
\usepackage[T1]{fontenc}    
\usepackage{hyperref}       
\usepackage{url}            
\usepackage{booktabs}       
\usepackage{amsfonts}       
\usepackage{nicefrac}       
\usepackage{microtype}      
\usepackage{xcolor}         
\usepackage{graphicx}
\usepackage{multirow} 
\usepackage{arydshln}
\usepackage{amssymb}
\usepackage[english]{babel}
\usepackage{blindtext}
\usepackage{subfigure}
\usepackage{wrapfig}
\usepackage{algpseudocode}
\usepackage{algorithmicx,algorithm}

\title{Variance-Aware Machine Translation Test Sets}

\author{Runzhe Zhan$^1$\thanks{~Equal contribution}~~~
        Xuebo Liu$^2\footnotemark[1]$~~~
        Derek F. Wong$^1$\thanks{~Corresponding author}~~~
        Lidia S. Chao$^1$ \\
  $^1$NLP$^2$CT Lab, Department of Computer and Information Science, 
  University of Macau \\
      \texttt{nlp2ct.runzhe@gmail.com, \{derekfw,lidiasc\}@um.edu.mo} \\
    $^2$Institute of Computing and Intelligence, Harbin Institute of Technology (Shenzhen) \\
          \texttt{liuxuebo@hit.edu.cn} }
          
\begin{document}

\maketitle

\begin{abstract}
We release 70 small and discriminative test sets for machine translation (MT) evaluation called \textit{variance-aware test sets} (VAT), covering 35 translation directions from WMT16 to WMT20 competitions. 
VAT is automatically created by a novel {\it variance-aware filtering method} that filters the indiscriminative test instances of the current MT test sets without any human labor.
Experimental results show that VAT outperforms the original WMT test sets in terms of the correlation with human judgement across mainstream language pairs and test sets.
Further analysis on the properties of VAT reveals the challenging linguistic features (e.g., translation of low-frequency words and proper nouns) for competitive MT systems, providing guidance for constructing future MT test sets.
The test sets and the code for preparing variance-aware MT test sets are freely available at \texttt{\href{https://github.com/NLP2CT/Variance-Aware-MT-Test-Sets}{https://github.com/NLP2CT/Variance-Aware-MT-Test-Sets}}.
\end{abstract}

\section{Introduction}
Automated machine translation (MT) evaluation relies on metrics and test sets. Based on the use of test sets, the metrics to quantify the performance of MT systems can be divided into two categories: reference-based metrics \citep{papineni-etal-2002-bleu, popovic-2015-chrf, lo-2019-yisi, zhang2019bertscore} and reference-free metrics \citep{popovic-2012-morpheme, yankovskaya-etal-2019-quality}. 
Reference-based metrics which measure the overlap between the reference and model's hypothesis, are widely used both in research and practice. Even the state-of-the-art metrics that exploit a pre-trained model \citep{zhang2019bertscore, sellam-etal-2020-bleurt, rei-etal-2020-comet} are able to evaluate the finer-grained semantic overlap, it still cannot achieve human-level judgements \citep{ma-etal-2019-results, mathur-etal-2020-results}.
Although the metric itself can be further elaborated, the reference in the test set, which is another key ingredient in the MT evaluation, has received less attention from the community.

The references are not innocent of confusing automatic metrics. Research has proven that the collected references tend to exhibit a monotonous translation style \citep{popovic-2019-reducing, freitag-etal-2020-bleu} instead of natural text, and lack diversity in the evaluation.
On the other hand, the competitive MT systems typically share a homogeneous architecture and training data, causing the performance of the MT systems to be too close to be distinguished, thus the differences in the scores given by the automatic metrics are small.
To alleviate this problem, previous work has focused on increasing the diversity of the references by means of paraphrasing, including human paraphrasing \citep{freitag-etal-2020-human, freitag-etal-2020-bleu} and automatic paraphrasing \citep{kauchak-barzilay-2006-paraphrasing, guo-hu-2019-meteor, bawden-etal-2020-study}, but both of these are expensive in terms of human labor and computational cost. 
Considering the fact that not all the references are monotonous, it is still unclear how to select those discriminative references instead of diversifying them for the MT evaluation.

This paper aims to tackle this problem without any human labor.
Our motivation comes from a common fact in the real world. In a general examination or test, the simplest and most difficult questions cannot tell the difference between the examinees because they may be all correctly or incorrectly answered. Accordingly, those questions that receive diverse answers play a vital role in distinguishing the examinees' abilities by comparing them with the ground truth. Based on this fact, a similar phenomenon can also happen in the test set for evaluating machine translations. 

In this paper, we use the variance of translation scores evaluated by the metric as a criterion to create a \textit{variance-aware} test set whose references are more discriminative in evaluating MT systems. 
The selected references are characterized by their diverse evaluation scores, indicating that the MT systems are not consistent in translating the same source, thus this translation case is a valuable indicator for distinguishing the capability of MT systems. 
Experimental results show that evaluating with the created \textit{variance-aware} test set can improve the correlation with human judgements. Further analysis of the properties of the \textit{variance-aware} test set also confirms its effectiveness.

Our main contributions are as follows:
\begin{itemize}
	\item We release 70 {\it variance-aware} MT test sets, covering 35 translation directions from the WMT16 to WMT20 competitions. The test set filters 60\% of the test instances from the original WMT version, which is time-efficient for research consuming high computational resources (e.g., reinforcement learning and neural architecture search). 
	\item We propose a simple and effective method to automatically identify discriminative test instances from MT test sets. We demonstrate that using the discriminative test instances can yield a better correlation with human judgements than using the original test set.
	\item We give an in-depth analysis of the properties of discriminative and non-discriminative test instances. We find that the translations of low-frequency words and proper nouns are highly discriminative, providing clues for building challenging MT test sets.

\end{itemize}

\section{Background}
\subsection{MT Evaluation and Meta-Evaluation}
The evaluation of machine translation is a crucial topic in the development of MT due to the need to compare the performance of several candidate MT systems.
Traditionally, human assessment is used to evaluate MT systems, but it is expensive in terms of its costs. Moreover, the quality of the assessment of crowdsourced evaluation work is unpredictable, and there is a big gap between non-expert and professional translators \citep{toral-etal-2018-attaining, laubli2020set, mathur-etal-2020-results}. 
Therefore, automatic evaluation metrics have received a lot of attention due to their advantages, such as their low cost and the controllability of the process, and are now widely used in model selection and optimization \citep{shen-etal-2016-minimum, wieting-etal-2019-beyond}. 

The reference-based metrics which rely on a reference translation are the most popular automatic evaluation metrics. They differ in the ways they measure overlap. For example, BLEU \citep{papineni-etal-2002-bleu} and its variants \citep{doddington2002automatic, popovic-2015-chrf} evaluate the overlap by matching the $n$-grams, and other metrics like TER \citep{Snover06astudy} quantify the overlap by the edit distance. However, these metrics are conducted in a hard matching paradigm and do not consider semantics. METEOR \citep{banerjee-lavie-2005-meteor, denkowski-lavie-2014-meteor} alleviates this problem by introducing synonymy and other linguistic features in the word matching but is limited in the availability of language resources. Recent embedding-based metrics break the limitation of hard matching, making it possible to evaluate the semantic overlap. By enhancing the semantic representation by a pre-trained model \citep{devlin-etal-2019-bert, lample2019cross}, BERTScore \citep{zhang2019bertscore} correlates better with human judgements than previous metrics. At the same time, the end-to-end paradigm using a pre-trained representation is also applied to the evaluation of MT, and has achieved remarkable performance, e.g., COMET \citep{rei-etal-2020-comet} and BLEURT \citep{sellam-etal-2020-bleurt}.

To verify the effectiveness of automatic evaluations, the process that measures the correlation between the scores given by an automatic metric and human ratings is called meta-evaluation. A meta-evaluation mainly uses correlation coefficients such as Pearson's $r$ to determine the extent to which the automatic metric performs like a human evaluator \citep{callison-burch-etal-2006-evaluating,callison2008further}. The validation process of our method covers the mainstream metrics and uses the ordinary meta-evaluation methods to validate the improvement in the correlation. 

\subsection{Shortcomings of Current Test Sets}
The less discriminative instances in a public benchmark are the bottleneck of automatic evaluation.
The test sets released by the organizers of the WMT competition are the well-recognized benchmark for MT evaluation. However, some have argued that some of the references in these test sets may mislead reference-based metrics, making the evaluation results of automatic metrics different from human judgement. 
One major issue is that the existing references tend to be monotonous \citep{popovic-2019-reducing, freitag-etal-2020-bleu}. This translation style is easy to achieve by the MT systems and less discriminative for the evaluation.
In addition, based on the phenomenon observed by \cite{zhan-etal-2021-difficulty} in the evaluation of the WMT19 English$\rightarrow$German task, most tokens can be correctly translated by all the participation systems, especially for the competitive ones, indicating that the test sets are only partly valuable in distinguishing the MT systems.

There are ways to create a more diverse test set so as to improve discernment. \cite{kauchak-barzilay-2006-paraphrasing} explored the automatic paraphrasing techniques for improving the accuracy of automatic metrics and validated their effectiveness on small-scale human assessment data. \cite{bawden-etal-2020-study} further investigated the use of automatic paraphrasing in automatic evaluation, finding limited gains in correlation with human judgements on the WMT19 benchmark. Promisingly, human paraphrased references have proved that they can significantly improve the correlation metrics of BLEU on some language pairs \citep{freitag-etal-2020-bleu}. Overall, these methods to augment the references are restricted by their construction costs and consistently limited improvement.

Instead of diversifying the references, our work pays attention to selecting the discriminative part from the existing test set for better distinguishing between strong MT systems.

\section{Variance-Aware Test Set}
\begin{figure}[h]
    \centering
    \includegraphics[width=0.98\textwidth]{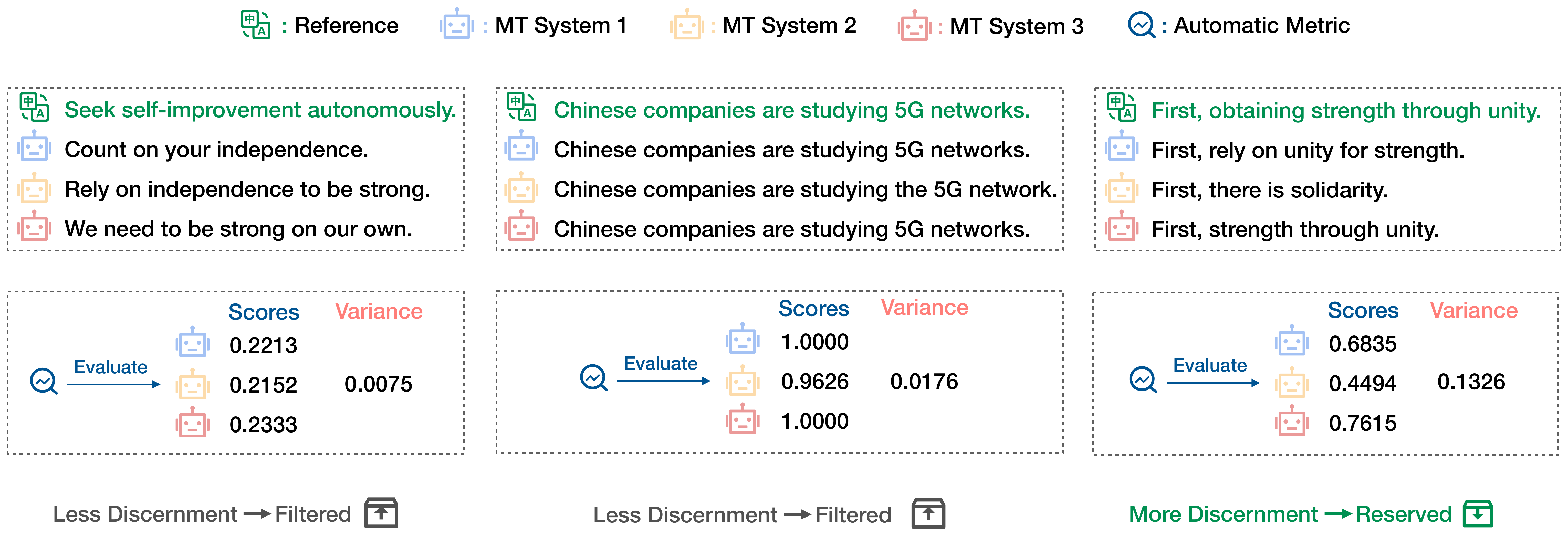} 
    \caption{An illustration of the proposed variance-aware filtering method.}
    \label{fig:illustrate}
\end{figure}
\subsection{Motivation}
Generally speaking, the test items that are either too easy or too difficult, cannot tell the differences of test-takers since they would not perform very differently when answering the extremely simple or difficult questions.

Machine translation evaluation is a kind of test. One can make an analogy between MT evaluation and tests in general: the MT systems are the test-takers, the instances in the test set are the test items. Similarly, to discriminate between MT systems in terms of their ability, a discriminative test instance must reveal clear differences in the systems' performance.
As illustrated in Figure \ref{fig:illustrate}, evaluating MT systems' performance by using the first two references causes a subtle difference in evaluation results due to the polarized difficulty, thus it is hard to discriminate between the systems in this circumstance. By contrast, the gap of evaluation results in the last case is huge enough to detect the differences in translation ability.

These cases clearly show that a discriminative test instance must make the evaluation exhibit a large diversity so that it can become a decisive clue for comparing the MT systems.
Since the metrics evaluate the performance of an MT system, the discrimination power of test instances can be quantified by the variance of scores given by a metric, reflecting the degree of diversity in the evaluation. A higher variance indicates that using this test instance in the evaluation makes it easier to differentiate between the systems.
Therefore, our goal is to create a discriminative test set for better evaluating MT systems by selecting the instances whose variance of evaluation results is high. This process will be referred to as \textit{variance-aware filtering}.

\subsection{Variance-Aware Filtering}
To measure how differently MT systems perform on a test instance, the performance of candidate systems is firstly quantified by the automatic metrics, then the standard deviation is simply used as a statistical indicator to model the diversity of the evaluated performance. The standard deviation takes the square root of the variance, we use it because the scale of this measurement is the same as the original data.
Given $N$ references $\mathcal{T}=\left\{t_1, t_2, ..., t_N \right\}$ and a set of corresponding hypotheses $\mathbf{h} =\left\{ \mathbf{h}_1, \mathbf{h}_2, ..., \mathbf{h}_N \right\}$ generated by $k$ systems in which $\mathbf{h}_i =\left\{ h^{(1)}_i, h^{(2)}_i, ..., h^{(k)}_i \right\}$, the performance diversity of hypothesis $\mathbf{h}_i$ is estimated by the standard deviation $\sigma_i$ of the scores, which is formulated as:

\begin{equation}
    {\sigma}_i=\sqrt{{\frac {1}{k}}\sum _{j=1}^{k}(\mathcal{M}(h^{(j)}_i, t_i)-{\mu_i})^{2}}, ~~ 1\leq i \leq N 
\end{equation}

where $\mathcal{M}(\cdot, \cdot)$ is the metric used to score the performance of the translation and $\mu_i$ is the average value of all the systems' scores, which can be calculated as follows:
\begin{equation}
    {\mu_i} = \frac{1}{k} \sum_{j=1}^{k} \mathcal{M}(h^{(j)}_i, t_i), ~~ 1\leq i \leq N 
\end{equation}

For all the standard deviations $\left\{\sigma_1, \sigma_2, ..., \sigma_N \right\}$, a higher $\sigma_i$ indicates that the behaviour of the systems on reference $t_i$ is more diverse. Therefore, $\lambda$ percent of test instances whose corresponding references have lower values of $\sigma$ will be filtered out in order to create a new discriminative test set, where $\lambda$ is a hyperparameter determined by the empirical experiments.

\section{Experiments and In-Depth Analysis}
\subsection{Experimental Setup}
\begin{table}[h]
\caption{Detailed information about the test sets involved in the experiments, where $\mathrm{Num}$ denotes the number of translation directions.}
\centering
\scalebox{0.8}{
\begin{tabular}{ccllll}
\toprule
\multicolumn{1}{c}{}             & \multicolumn{1}{c}{\begin{tabular}[c]{@{}c@{}}\textbf{WMT16}\\ ($\mathrm{Num}$=7)\end{tabular}}   & \multicolumn{1}{c}{\begin{tabular}[c]{@{}c@{}}\textbf{WMT17}\\ ($\mathrm{Num}$=14)\end{tabular}}           & \multicolumn{1}{c}{\begin{tabular}[c]{@{}c@{}}\textbf{WMT18}\\ ($\mathrm{Num}$=14)\end{tabular}}           & \multicolumn{1}{c}{\begin{tabular}[c]{@{}c@{}}\textbf{WMT19}\\ ($\mathrm{Num}$=18)\end{tabular}}  & \multicolumn{1}{c}{\begin{tabular}[c]{@{}c@{}}\textbf{WMT20}\\ ($\mathrm{Num}$=17)\end{tabular}}                                 \\
\midrule
\textbf{X-English}   & \begin{tabular}[c]{@{}l@{}}cs, de, fi, ro, ru, \\tr\end{tabular} & \begin{tabular}[c]{@{}l@{}}cs, de, fi, lv, ru, \\tr, zh\end{tabular} & \begin{tabular}[c]{@{}l@{}}cs, de, et, fi, ru,\\ tr, zh\end{tabular} & \begin{tabular}[c]{@{}l@{}}de, fi, gu, kk, lt,\\ru, zh\end{tabular}     & \begin{tabular}[c]{@{}l@{}}cs, de, iu, ja, km,\\ pl, ps, ru, ta, zh\end{tabular} \\\hdashline
\textbf{English-X}   & ru                                                                & \begin{tabular}[c]{@{}l@{}}cs, de, fi, lv, ru,\\tr, zh\end{tabular} & \begin{tabular}[c]{@{}l@{}}cs, de, et, fi, ru,\\tr, zh\end{tabular} & \begin{tabular}[c]{@{}l@{}}cs, de, fi, gu, kk,\\lt, ru, zh\end{tabular} & \begin{tabular}[c]{@{}l@{}}cs, de, ja, pl, ru, \\ ta, zh\end{tabular}  \\ \hdashline          
\textbf{Others} & \multicolumn{1}{c}{/} & \multicolumn{1}{c}{/} & \multicolumn{1}{c}{/} & de-cs, de-fr, fr-de & \multicolumn{1}{c}{/} \\
\bottomrule
\end{tabular}
}
\label{tab:tl}
\end{table}
\paragraph{Data} Five WMT test sets \citep{bojar-etal-2016-results, bojar-etal-2017-results, ma-etal-2018-results, ma-etal-2019-results,mathur-etal-2020-results} ranging from WMT16 to WMT20 were used to conduct the experiments since they are the well-recognized benchmarks in the MT community. The included translation directions are as shown in Table \ref{tab:tl}. 
On the other hand, we choose the test sets starting from WMT16 because the neural machine translation \citep{DBLP:journals/corr/BahdanauCB14, sennrich-etal-2016-neural, DBLP:conf/nips/VaswaniSPUJGKP17} paradigm has largely improved the capability of MT systems and the systems submitted to the WMT competitions have gradually become more competitive since 2016. 
For each language pair, we use the raw data released by the WMT competitions including the official references, submitted hypotheses of the different MT systems, and the corresponding human ratings. 

\paragraph{Metrics and Meta-Evaluation}
Without loss of generality, we validate our research hypotheses on the following four representative metrics, and use their public open-source implementations so that the results can be easily reproduced:
\begin{itemize}
    \item \textbf{BLEU} \citep{papineni-etal-2002-bleu} is an $n$-gram based metric that uses the precision rate to evaluate the coverage of reference $n$-gram in the model hypothesis. We use the sentence-level BLEU in the filtering procedure and evaluate the corpus-level system performance.
    \item \textbf{COMET} \citep{rei-etal-2020-comet} is an end-to-end metric that builds on the top of the pre-trained XLM model including reference-based models and reference-free models. We use the recommended reference-based estimator model in the experiments.
    \item \textbf{BLEURT} \citep{sellam-etal-2020-bleurt} is an end-to-end metric that fine-tunes the BERT model with several regression and classification tasks to make the model better be adapted to the MT evaluation scenario. We use its released checkpoint and default settings in the experiments.
    \item \textbf{BERTScore} \citep{zhang2019bertscore} is an embedding-based metric that relies on a pre-trained BERT model to encode the reference and hypothesis, measuring the similarity of representation with precision (BERTS-P), recall (BERTS-R), and the $F$-measure (BERTS-F). For the evaluation of different language pairs, we use the default BERT-family models as the same as the BERTScore implementation, e.g., RoBERTa-large for evaluating English text and BERT-base-multilingual-case for evaluating other languages.
\end{itemize}
To examine the effectiveness of the automatic evaluation metrics, we use the system-level Pearson's $r$, Kendall's $\tau$ and Spearman's $\rho$ correlation coefficients as the metrics for measuring how the results of the automated evaluation correlate with human judgements; these are also widely used in the competitions \citep{machacek-bojar-2013-results, mathur-etal-2020-results} and related research \citep{freitag-etal-2020-bleu}.

\subsection{Ablation Study} \label{section:abl}
There are two main factors that may affect our proposed filtering approach: the filtering percentage $\lambda$ and the filtering metric $\mathcal{M}$. Hence, a series of empirical experiments was conducted on the WMT20 benchmark to explore the best settings for building the most discriminative test sets, and the finalized setting would subsequently be used to validate its generality in other WMT benchmarks.

\begin{figure}[h]
    \centering
    \subfigure[BLEU and End-to-End Metrics]{
    \includegraphics[scale=0.6]{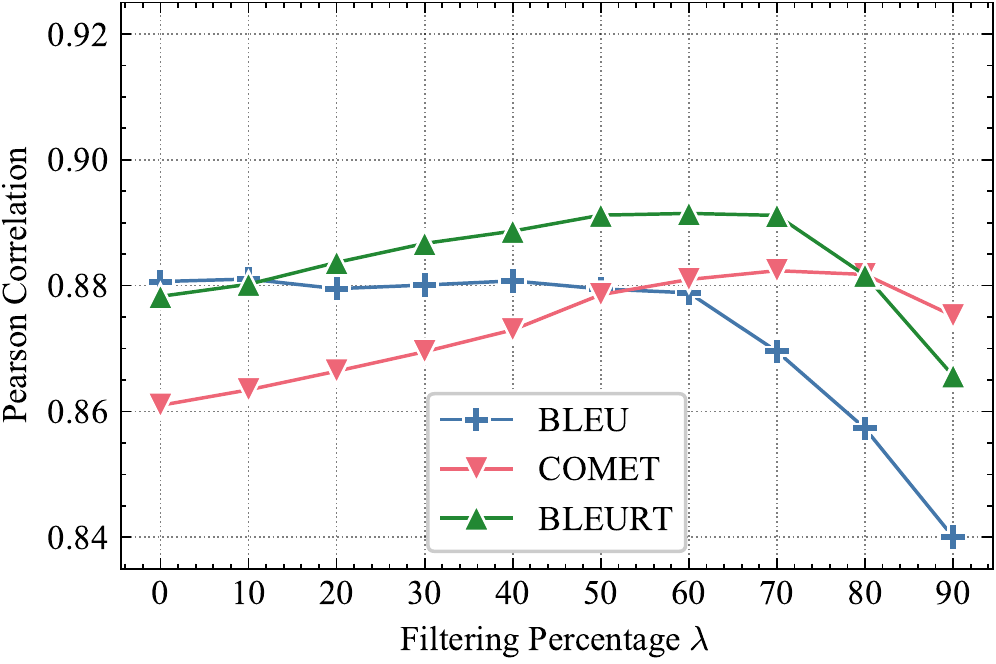} 
    }
    \subfigure[BERTScore Metrics]{
    \includegraphics[scale=0.6]{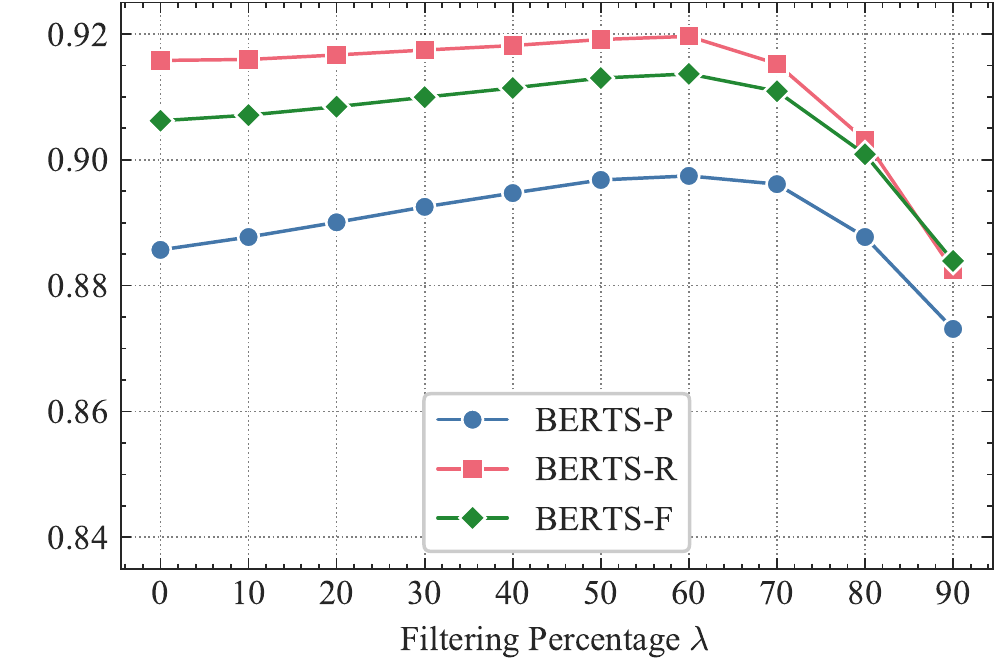}
    }
    \caption{Comparison of averaged Pearson correlations measured on all the WMT20 translation directions using different filtering percentages. Extreme settings will hurt the correlation results whereas filtering 60\%\ or 70\% of the data out of the test sets is the appropriate choice.}
    \label{fig:percecntage}
\end{figure}
\begin{figure}[h]
    \centering
    \subfigure[X-English]{
    \includegraphics[scale=0.6]{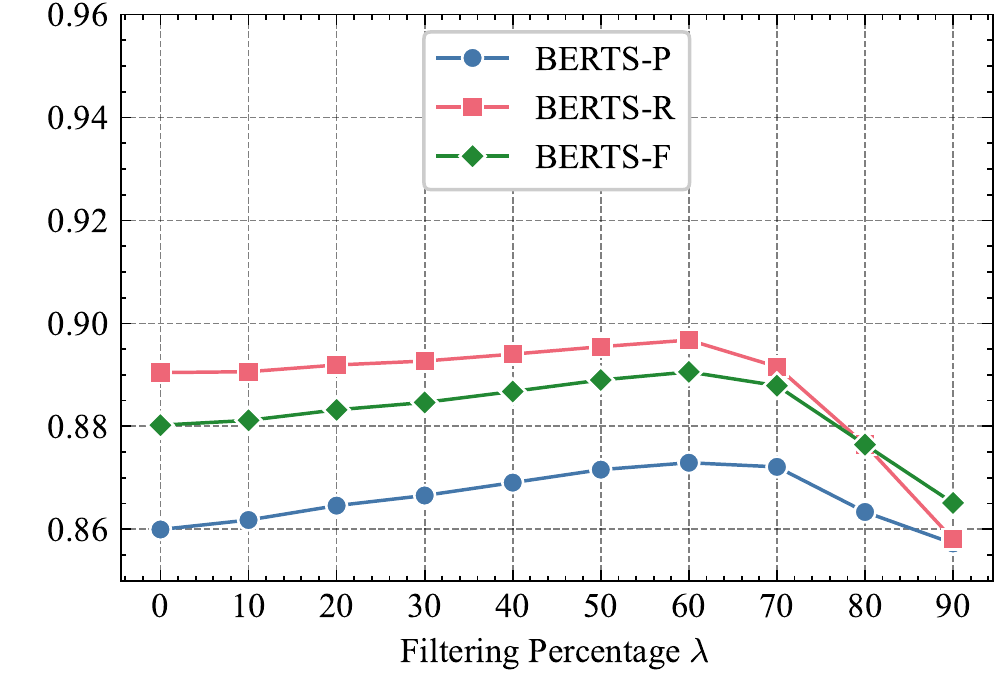} 
    }
    \subfigure[English-X]{
    \includegraphics[scale=0.6]{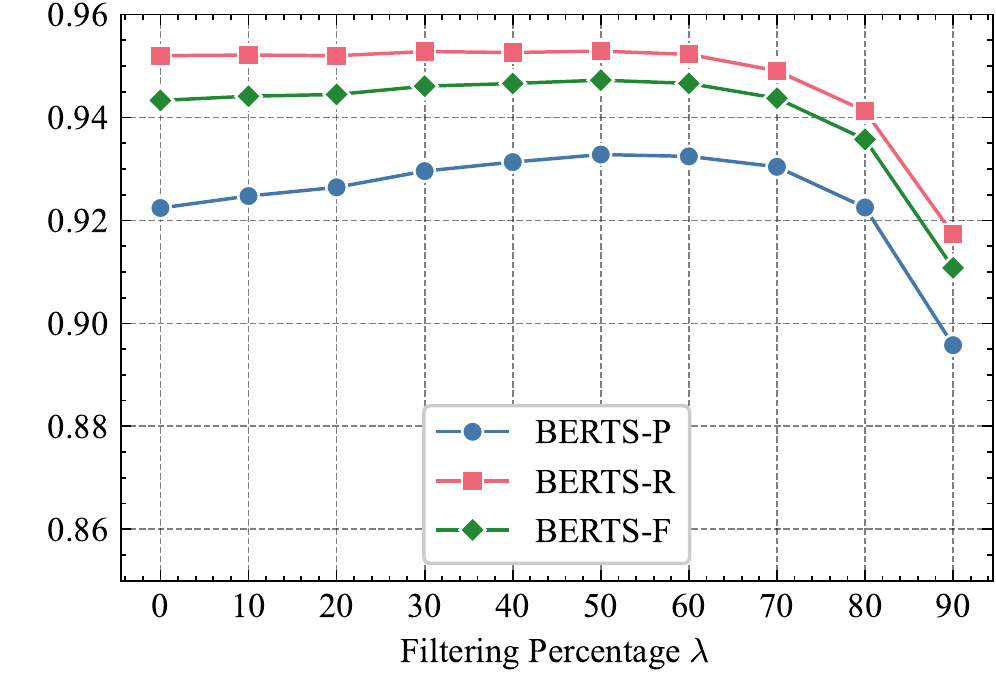}
    }
    \caption{Comparison of averaged Pearson correlations measured on the WMT20 to/from English translation directions using different filtering percentages. Filtering 60\% of original data works well for the two translation directions.}
    \label{fig:languages}
\end{figure}

\paragraph{Choice of Filtering Percentage $\lambda$} affects the amount of data to be preserved and is also an indicator that reflects the discernment of the current data sets in terms of the evaluation metrics.
As shown in Figure \ref{fig:percecntage}, using only a partial test set can improve the evaluation correlation of automatic metrics, but the most effective percentage setting depends on the type of evaluation metric.
Compared to the BLEU metric, the metrics driven by the pre-trained models achieve the local optimal correlation using a smaller proportion of the test set, i.e., $\lambda \geq 50$.
The underlying reason for this may lie in the granularity of the evaluation in terms of the semantics: a metric that is better in parsing the semantics needs fewer data to distinguish the MT systems because of the larger impact of those discriminative samples in the comparison.
However, the percentage setting may vary from language to language. 
Figure \ref{fig:languages} shows that filtering 60\% of original data still can improve the correlation performance for both to-English and from-English translation directions, confirming the robustness of this setting.
For fitting most of the metrics and languages, we filter 60\% of the instances out of the original test sets in the subsequent experiments.

\begin{figure}[h]
    \centering
    \subfigure[Pearson's $r$] {
    \includegraphics[scale=0.55]{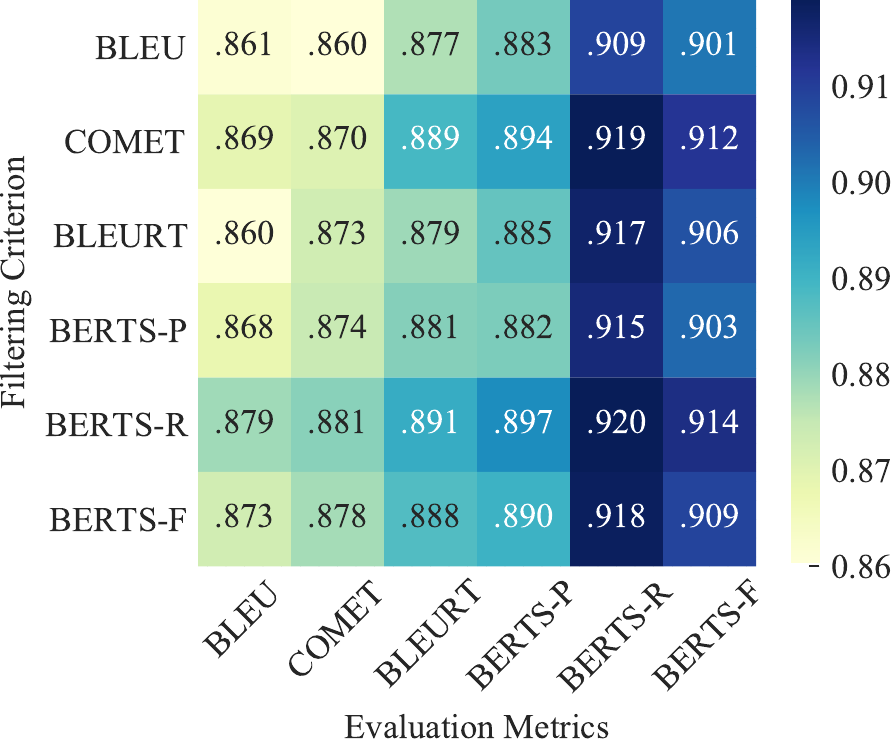} 
    }
    \subfigure[Kendall's $\tau$]{
    \includegraphics[scale=0.55]{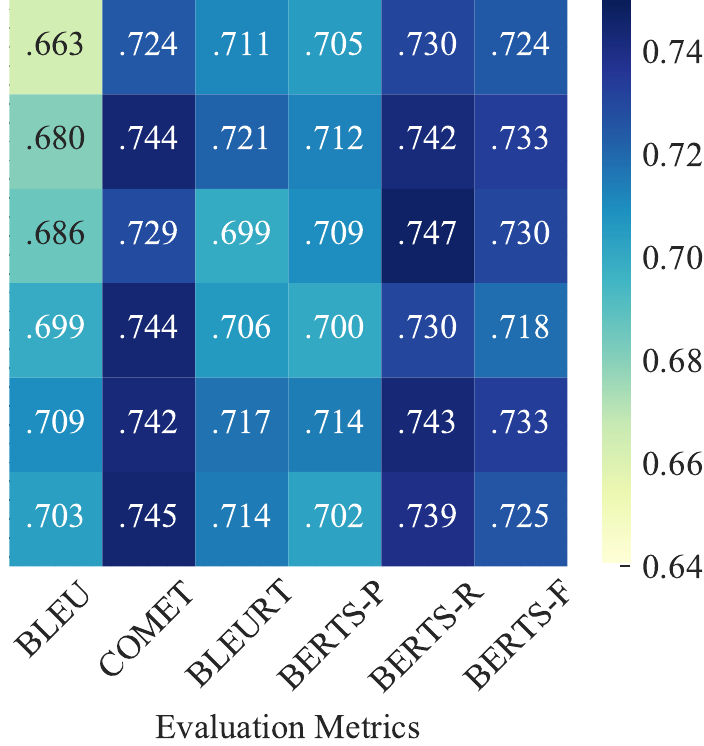} 
    }
    \subfigure[Spearman's $\rho$]{
    \includegraphics[scale=0.55]{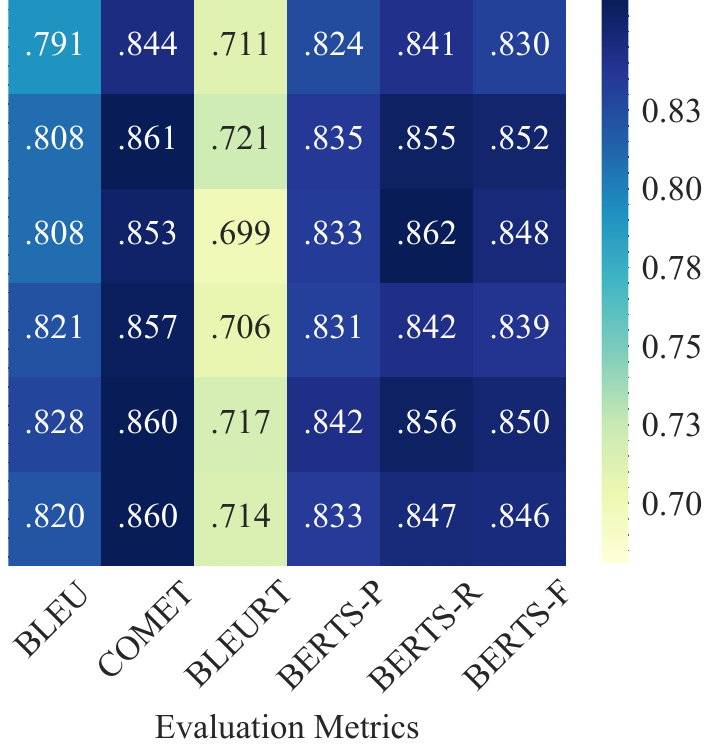} 
    }
    \caption{Comparison of averaged correlation results measured on all the WMT20 translation directions using different filtering metrics. Filtering the test sets by BERTS-R scores consistently yields stable correlation results across different evaluation metrics. Using COMET scores has comparable correlation performance with BERTR-S except Pearson correlation results.}
    \label{fig:filterby}
\end{figure}

\paragraph{Choice of the Filtering Metric $\mathcal{M}$} matters because the discernment of a test instance can not be estimated without an accurate evaluation of the MT systems' performance.
Figure \ref{fig:filterby} presents how the scores given by the different metrics affect the correlation of filtered test sets. 
Filtering the test set based on the scores given by the BERTS-R metric outperforms the test sets created by other metrics. It is reasonable that the BERTS-R metric consistently achieves the best correlation when using it as the evaluation metric (also as shown in Figure \ref{fig:percecntage}), and thus is better at quantifying the differences between the hypotheses and filtering out the non-discriminative instances.
Although COMET is also remarkable in terms of Kendall's $\tau$ correlation, we chose to use the scores given by the BERTS-R metric as the bedrock to create the discriminative test set due to the balancing performance of these correlation coefficients. 

\subsection{Main Results} \label{section:res}
Using the filtering settings determined in the previous sections for other WMT benchmarks, Tables \ref{table:corr_main} and \ref{table:corr_lang} present the comparison of correlation results between using the filtered and original test sets. The improved correlation performance across most metrics and benchmarks consistently confirms the greater effectiveness of evaluating with a variance-aware test set (VAT), especially for the metrics powered by pre-trained models.
As for the $n$-gram-based metrics, it may over-penalize overlaps that share the same semantics, due to the hard-matching paradigm, making some VAT instances inactive in evaluating the diverse hypotheses. 
In contrast to the hard-matching paradigm, the metrics using the pre-trained models are able to fairly judge synonymous expressions, thus the created VAT are substantially useful to distinguish the MT systems.

\begin{table}[h]
\centering
\caption{Comparison of averaged correlation results using original and variance-aware test sets (\textit{VAT}) where $\mathrm{Num}$ denotes the number of language pairs. Evaluating MT systems with variance-aware test sets (\textit{+VAT}) better correlates with human judgements across different evaluation metrics.}
\scalebox{0.8}{
\begin{tabular}{rccccccccccccccc}
\toprule
\multicolumn{1}{r}{\multirow{2}{*}{\textbf{Metric}}} & \multicolumn{3}{c}{\begin{tabular}[c]{@{}c@{}}\textbf{WMT16}\\ ($\mathrm{Num}$=7)\end{tabular}} & \multicolumn{3}{c}{\begin{tabular}[c]{@{}c@{}}\textbf{WMT17}\\ ($\mathrm{Num}$=14)\end{tabular}} & \multicolumn{3}{c}{\begin{tabular}[c]{@{}c@{}}\textbf{WMT18}\\ ($\mathrm{Num}$=14)\end{tabular}} & \multicolumn{3}{c}{\begin{tabular}[c]{@{}c@{}}\textbf{WMT19}\\ ($\mathrm{Num}$=18)\end{tabular}} & \multicolumn{3}{c}{\begin{tabular}[c]{@{}c@{}}\textbf{WMT20}\\ ($\mathrm{Num}$=17)\end{tabular}} \\
\cmidrule(r){2-4} \cmidrule(r){5-7} \cmidrule(r){8-10} \cmidrule(r){11-13} \cmidrule(r){14-16}
              & $|r|$         & $|\tau|$   & $|\rho|$    & $|r|$         & $|\tau|$   & $|\rho|$    & $|r|$         & $|\tau|$   & $|\rho|$    & $|r|$         & $|\tau|$   & $|\rho|$    & $|r|$         & $|\tau|$   & $|\rho|$    \\
\midrule
\textbf{BLEU}          & .826          & .645          & .778          & .910          & .737          & .865          & .827 & \textbf{.727} & \textbf{.802} & .912          & .762          & .878          & \textbf{.881} & .675          & .798          \\
\textit{\textbf{+VAT}} & \textbf{.880} & \textbf{.723} & \textbf{.837} & \textbf{.928} & \textbf{.754} & \textbf{.876} & \bf .827          & .723          & .796          & \textbf{.918} & \textbf{.786} & \textbf{.906} & .879          & \textbf{.709} & \textbf{.828} \\ \hdashline
\textbf{COMET}         & .988          & \textbf{.886} & \textbf{.958} & .982          & .884          & .958          & .980          & \textbf{.925} & \textbf{.975} & .979          & \textbf{.882} & \textbf{.958} & .861          & .731          & .854          \\
\textit{\textbf{+VAT}} & \textbf{.988} & .881          & .955          & \textbf{.985} & \textbf{.885} & \textbf{.959} & \textbf{.983} & .921          & .969          & \textbf{.981} & .862          & .948          & \textbf{.881} & \textbf{.743} & \textbf{.860} \\ \hdashline
\textbf{BLEURT}        & .982          & .856          & .942          & .939          & .789          & .890          & .970          & \textbf{.900} & \textbf{.966} & .925          & .776          & .896          & .878          & .699          & .828          \\
\textit{\textbf{+VAT}} & \textbf{.984} & \textbf{.859} & \textbf{.944} & \textbf{.951} & \textbf{.807} & \textbf{.906} & \textbf{.974} & .891          & .959          & \textbf{.935} & \textbf{.791} & \textbf{.902} & \textbf{.892} & \textbf{.717} & \textbf{.839} \\ \hdashline
\textbf{BERTS-P}       & .970          & .848          & .924          & .951          & .806          & .909          & .965          & \textbf{.866} & \textbf{.949} & .953          & .811          & .911          & .886          & .699          & .827          \\
\textit{\textbf{+VAT}} & \textbf{.976} & \textbf{.880} & \textbf{.948} & \textbf{.960} & \textbf{.820} & \textbf{.919} & \textbf{.978} & .865          & .949          & \textbf{.953} & \textbf{.827} & \textbf{.924} & \textbf{.897} & \textbf{.714} & \textbf{.842} \\ \hdashline
\textbf{BERTS-R}       & .941          & .831          & .931          & \textbf{.974} & .825          & \textbf{.926} & .915          & \textbf{.843} & .908          & \textbf{.961} & .821          & .924          & .916          & .742          & .853          \\
\textit{\textbf{+VAT}} & \textbf{.953} & \textbf{.854} & \textbf{.943} & .972          & \textbf{.826} & .925          & \textbf{.953} & .842          & \textbf{.911} & .960          & \textbf{.834} & \textbf{.930} & \textbf{.920} & \textbf{.743} & \textbf{.856} \\ \hdashline
\textbf{BERTS-F}       & .975          & .881          & .950          & .970          & .833          & .927          & .947          & .846          & .909          & \textbf{.963} & \textbf{.824} & .924          & .906          & .728          & .848          \\
\textit{\textbf{+VAT}} & \textbf{.979} & \textbf{.900} & \textbf{.964} & \textbf{.974} & \textbf{.842} & \textbf{.929} & \textbf{.969} & \textbf{.873} & \textbf{.942} & .960          & .823          & \textbf{.925} & \textbf{.914} & \textbf{.733} & \textbf{.850}
\\ \bottomrule
\end{tabular}
}
\label{table:corr_main}
\end{table}

\begin{table}[h]
\centering
\caption{Comparison of Pearson correlations using original and variance-aware test sets (\textit{VAT}) on some mainstream language pairs. \textit{T.} denotes the WMT test set. Using variance-aware test sets (\textit{+VAT}) consistently improves the evaluation results of the language pairs across different test sets.}
\scalebox{0.8}{
\begin{tabular}{rlllllllllllllll}
\toprule
\multicolumn{1}{r}{\multirow{2}{*}{\textbf{Metric}}}   & \multicolumn{3}{c}{\textbf{De-En}}                                                      & \multicolumn{3}{c}{\textbf{En-De}}                                                      & \multicolumn{3}{c}{\textbf{Zh-En}}                                                      & \multicolumn{3}{c}{\textbf{En-Zh}}                                                      & \multicolumn{3}{c}{\textbf{En-Cs}}                                             \\
 \cmidrule(r){2-4} \cmidrule(r){5-7} \cmidrule(r){8-10} \cmidrule(r){11-13} \cmidrule(r){14-16} 
 & \textit{T.17}         & \textit{T.18}   & \textit{T.19}    & \textit{T.17}         & \textit{T.18}   & \textit{T.19}    & \textit{T.17}         & \textit{T.18}   & \textit{T.19}    & \textit{T.17}         & \textit{T.18}   & \textit{T.19}    & \textit{T.17}         & \textit{T.18}   & \textit{T.19}    \\
 \midrule
\textbf{BLEU}          & .928                     & .969                     & .888                     & .819                     & .980                     & .952                     & .869                     & .983                     & .900                     & .980                     & .947                     & .902                     & .956                     & \textbf{.996}            & .987                     \\
\textit{\textbf{+VAT}} & \textbf{.940}            & \textbf{.975}            & \textbf{.925}            & \textbf{.845}            & \textbf{.981}            & \textbf{.952}            & \textbf{.894}            & \textbf{.986}            & \textbf{.895}            & \textbf{.980}            & \textbf{.953}            & \textbf{.925}            & \textbf{.961}            & .994                     & \textbf{.994}            \\ \hdashline
\textbf{COMET}         & .989                     & .997                     & .947                     & .935                     & .989                     & .987                     & .979                     & .988                     & .989                     & \textbf{.993}            & .981                     & .975                     & .978                     & .974                     & .970                     \\
\textit{\textbf{+VAT}} & \textbf{.993}            & \textbf{.998}            & \textbf{.952}            & \textbf{.950}            & \textbf{.990}            & \textbf{.993}            & \textbf{.979}            & \textbf{.990}            & \textbf{.992}            & .990                     & \textbf{.985}            & \textbf{.976}            & \textbf{.985}            & \textbf{.976}            & \textbf{.978}            \\\hdashline
\textbf{BLEURT}        & .965                     & .997                     & .940                     & .797                     & .987                     & \textbf{.982}            & .915                     & .984                     & .984                     & .797                     & .883                     & .807                     & .919                     & \textbf{.990}            & \textbf{.987}            \\
\textit{\textbf{+VAT}} & \textbf{.979}            & \textbf{.998}            & \textbf{.944}            & \textbf{.841}            & \textbf{.987}            & .981                     & \textbf{.955}            & \textbf{.988}            & \textbf{.984}            & \textbf{.822}            & \textbf{.914}            & \textbf{.877}            & \textbf{.947}            & .986                     & .984                     \\\hdashline
\textbf{BERTS-P}       & .948                     & .998                     & .947                     & .798                     & .988                     & .984                     & .964                     & .981                     & .975                     & .970                     & .954                     & .881                     & .959                     & .994                     & .975                     \\
\textit{\textbf{+VAT}} & \textbf{.964}            & \textbf{.999}            & \textbf{.952}            & \textbf{.830}            & \textbf{.989}            & \textbf{.989}            & \textbf{.977}            & \textbf{.984}            & \textbf{.982}            & \textbf{.982}            & \textbf{.959}            & \textbf{.926}            & \textbf{.968}            & \textbf{.998}            & \textbf{.984}            \\\hdashline
\textbf{BERTS-R}       & .988                     & .997                     & .946                     & .909                     & .990                     & .991                     & \textbf{.981}            & .990                     & .987                     & \textbf{.994}            & .976                     & .940                     & .982                     & .997                     & .984                     \\
\textit{\textbf{+VAT}} & \textbf{.989}            & \textbf{.997}            & \textbf{.950}            & \textbf{.912}            & \textbf{.990}            & \textbf{.991}            & .978                     & \textbf{.991}            & \textbf{.987}            & .988                     & \textbf{.980}            & \textbf{.951}            & \textbf{.984}            & \textbf{.997}            & \textbf{.989}            \\\hdashline
\textbf{BERTS-F}       & .973                     & .999                     & .949                     & .859                     & \textbf{.990}            & .990                     & .983                     & .988                     & .983                     & .992                     & .968                     & .925                     & .976                     & .997                     & .981                     \\
\textit{\textbf{+VAT}} & \textbf{.981}            & \textbf{.999}            & \textbf{.952}            & \textbf{.876}            & .989                     & \textbf{.992}            & \textbf{.988}            & \textbf{.990}            & \textbf{.986}            & \textbf{.994}            & \textbf{.972}            & \textbf{.949}            & \textbf{.979}            & \textbf{.998}            & \textbf{.987}           
\\ \bottomrule
\end{tabular}
}
\label{table:corr_lang}
\end{table}

\subsection{Analysis of Variance-Aware Test Sets} \label{section:ana}
To investigate how the correlation improvement benefits from VAT, we characterize the VAT built on the WMT20 benchmark from the perspective of their linguistic and data properties in this section.

\begin{figure}[h]
    \centering
    \includegraphics[width=0.9\textwidth]{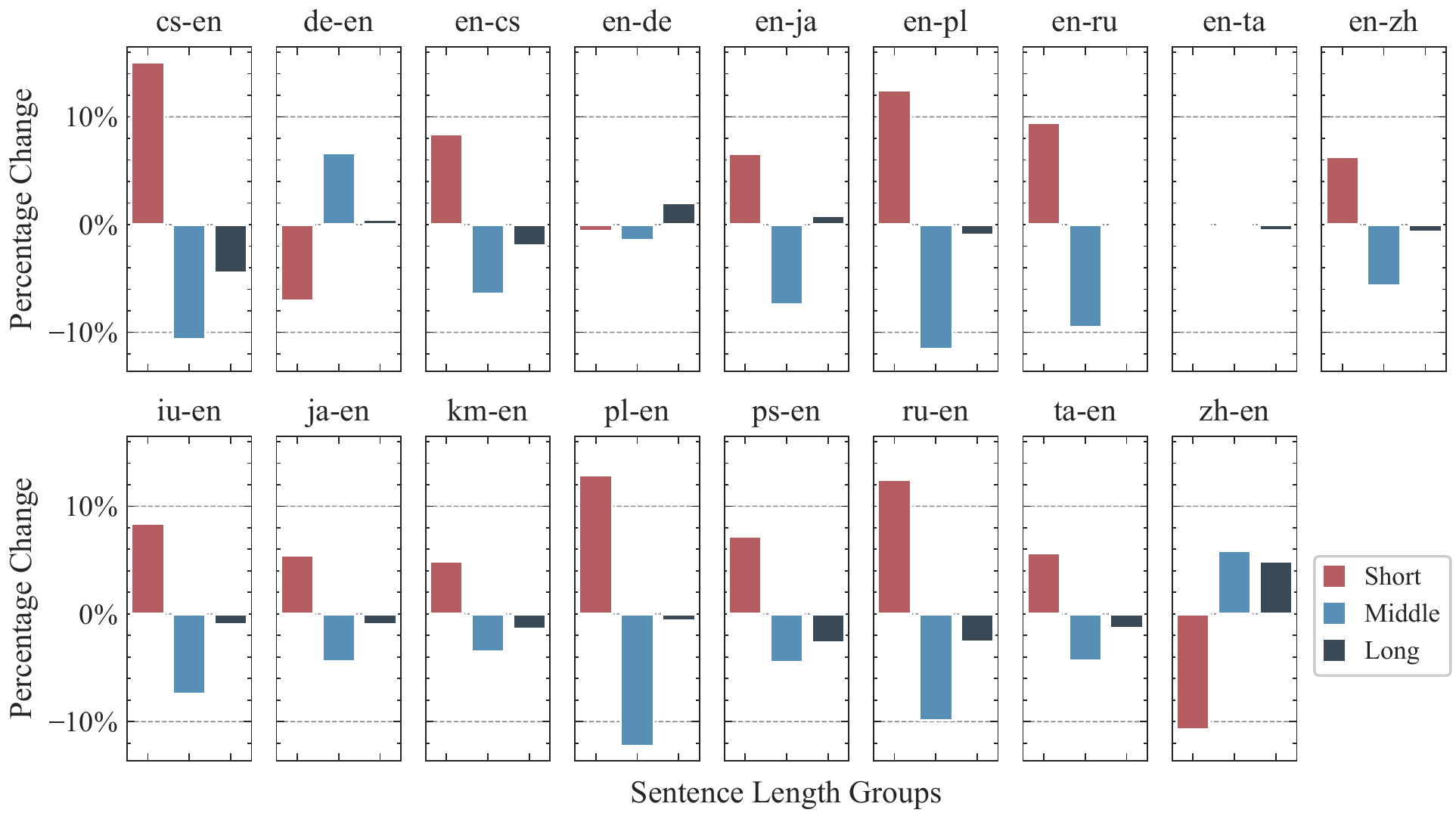} 
    \caption{Absolute constitution changes of variance-aware test sets in terms of sentence length. VAT preserved more short sentences for most language pairs. The one-third of the sentences will be treated as the Long group, and the Medium/Short grouping methods are analogous.}
    \label{fig:sent-length}
\end{figure}
\paragraph{Sentence Length} generally associates with the translation difficulty \citep{koehn-knowles-2017-six}, but the difficult sentence may be less relevant to the high discernment.
As shown in Figure \ref{fig:sent-length}, longer sentences exhibit lower discernment and were filtered out by our method.
Translating longer sentences is extremely challenging for MT systems due to the long-distance dependency or complex entity relationships \citep{cho-etal-2014-properties, sennrich-haddow-2016-linguistic, eriguchi2019incorporating}, leading to close translation performance of MT systems.
On the contrary, short sentences are more discriminative because different systems tend to show greater differences in their syntactic and lexical choices.
But some special translation directions show the opposite trends, such as Chinese$\rightarrow$English and German$\rightarrow$English. 
Not only is the number of systems that participated in these translation tasks relatively huge, but also the systems trained on these high-resource language pairs are likely to be more competitive. 
Since the competitive systems can be good at translating short sentences, the clues to judge their capability could rely on the translation of medium or long sentences.

\begin{figure}[h]
    \centering
    \includegraphics[width=0.9\textwidth]{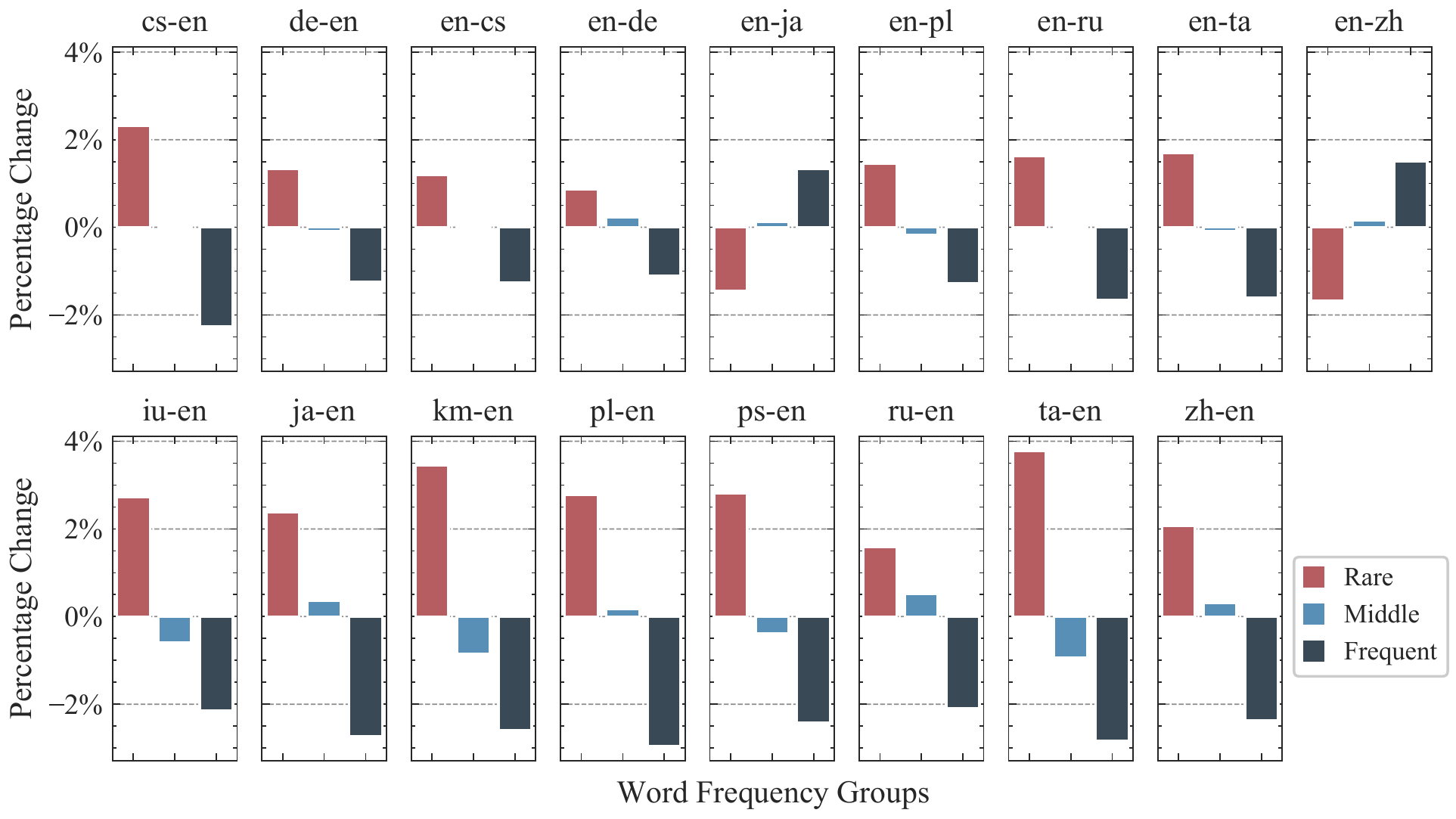}
    \caption{Absolute constitution changes of variance-aware test sets in terms of word frequency. VAT filtered the sentences which contain more frequent words. The boundaries for categorizing the ``Rare'', ``Middle'', ``Frequent'' group are 20\%, 60\%, 100\% percentile of word frequency, respectively.}
    \label{fig:frequency}
\end{figure}
\paragraph{Word Frequency} is a measure that reflects the finer-grained difference of MT systems since they may vary in their lexical choice of rare words \citep{koehn-knowles-2017-six, ding2021understanding}.
As shown in Figure \ref{fig:frequency}, the proportion of frequently occurring words in the training set is reduced in the VAT, indicating that high-frequency words are less discriminative.
The representations of high-frequency words learned on the training set tend to be stable, whereas the low-frequency words are insufficiently learned. In particular, some systems may enhance the translation performance of low-frequency words with the help of data augmentation \citep{fadaee-etal-2017-data, ding2021rejuvenation} or representation enhancement techniques \citep{ nguyen-chiang-2018-improving, liu-etal-2019-shared}, resulting in the differences of lexical choice performed on the test set.
Overall, the percentage change of word frequencies is not so large as it was for the comparison of sentence lengths, because the filtering operation is conducted at the sentence level, thus only those sentences whose proportion of low-frequency words is high will be preserved.

\begin{figure}[h]
    \centering
    \includegraphics[width=0.9\textwidth]{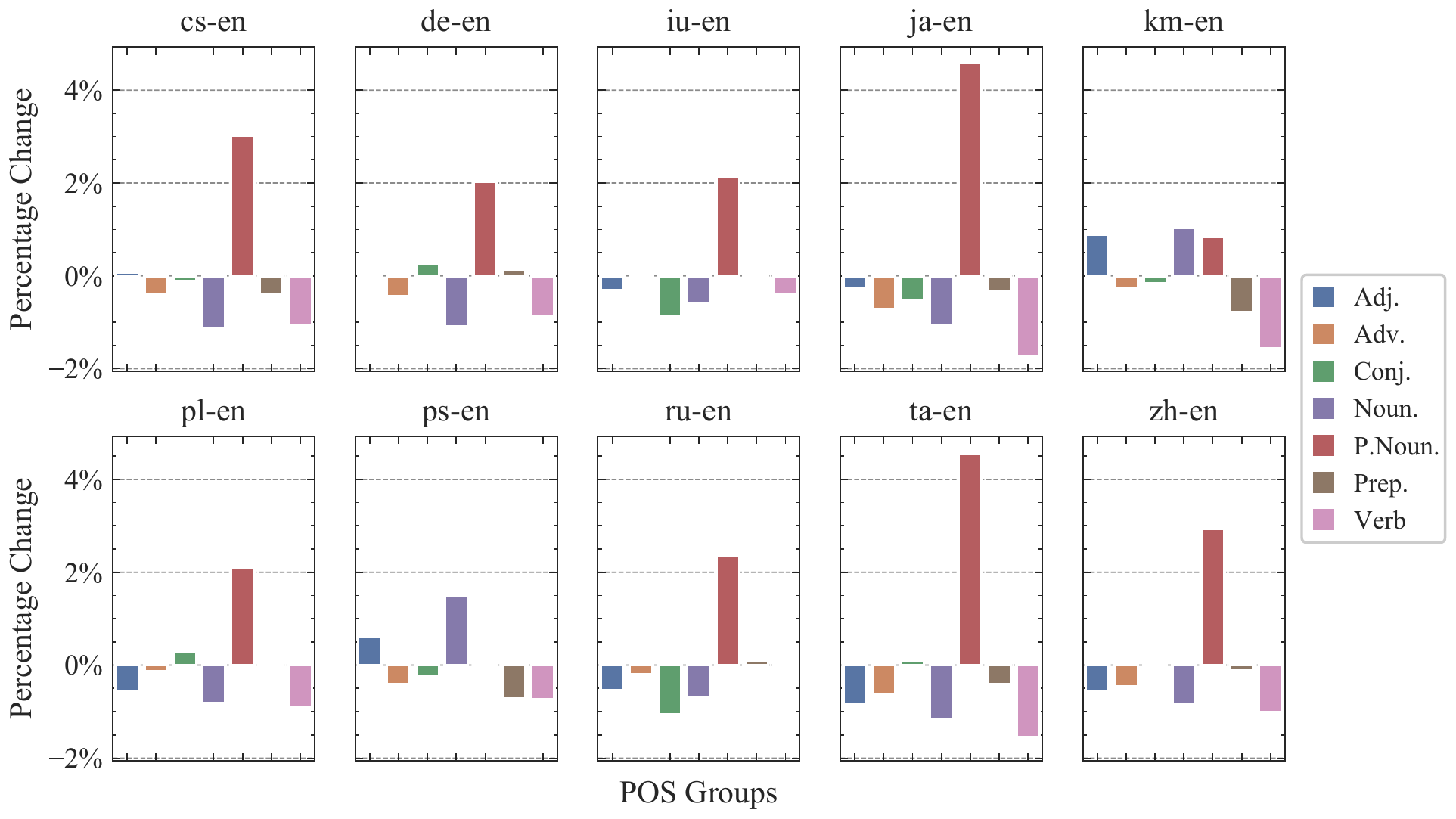} 
    \caption{Absolute constitution changes of variance-aware test sets in terms of English part-of-speech tagged by the \textsc{nltk} toolkit. VAT has more proper nouns than the original test sets.}
    \label{fig:pos}
\end{figure}

\paragraph{Part-of-Speech} better depicts the lexical features of VAT considering the syntactic role a word plays.
It can be seen from Figure \ref{fig:pos} that VAT preserved more sentences containing proper nouns (NNP).
This phenomenon echoes our previous comparative exploration of word frequency since there is a large overlap between NNPs and low-frequency words, such as the technical terms of a specific domain, but the translation performance on NNPs is not as intractable on long sentences. 
Due to the fact that the bottleneck of long-sentence translation may be related to the model architecture \citep{cho-etal-2014-properties}, most MT systems that share a homogeneous architecture \citep{DBLP:conf/nips/VaswaniSPUJGKP17} still have problems in translating these challenging sentences.
Similar to the problem of low-frequency words, the poor translation accuracy for NNPs can be alleviated by introducing external knowledge \citep{chatterjee-etal-2017-guiding} or domain adaptation techniques \citep{hu-etal-2019-domain-adaptation} concerning data-efficient learning, thus evaluating the translation of NNPs is also valuable in distinguishing MT systems.

\paragraph{Human Paraphrasing}
is another option for optimizing existing MT test sets. It asks human experts to paraphrase the references as much as possible~\citep{freitag-etal-2020-bleu}.
This can effectively improve the correlation of automatic metrics but is very costly.
In this experiment, we investigate the relationship between human paraphrased test sets provided by \cite{freitag-etal-2020-bleu} and the preserved (discriminative) and filtered out (non-discriminative) subsets by our approach.
\begin{wraptable}{r}{0.5\textwidth}
\vspace{-18pt}
\centering
\caption{Edit distances between different subsets of the English$\rightarrow$German test set with corresponding human paraphrased data. Human experts need to do fewer paraphrases on the preserved subsets.}
\begin{tabular}{cccc}
\toprule
               & \textbf{All} & \textbf{Filtered Out} & \textbf{Preserved} \\
               \midrule
\textbf{WMT20} & 30.35       & 30.46            & \textbf{30.18}   \\
\textbf{WMT19} & 19.82       & 20.25            & \textbf{19.17}   \\
\textbf{WMT18} & 20.02       & 20.30            & \textbf{19.72}  \\
\bottomrule
\end{tabular}
\label{table:edit_distance}
\vspace{-15pt}
\end{wraptable}
Table \ref{table:edit_distance} gives the averaged edit distance \citep{levenshtein1966binary}: a large value means more paraphrases on a subset.
The results show that human experts need to produce more paraphrases for the filtered out test instances whereas making fewer paraphrases for the preserved ones.
This means that the preserved subsets are of higher quality for MT evaluation, and thus can improve the correlation with human judgements.

\begin{wrapfigure}{r}{0.4\textwidth}
    \centering
    \includegraphics[width=0.4\textwidth]{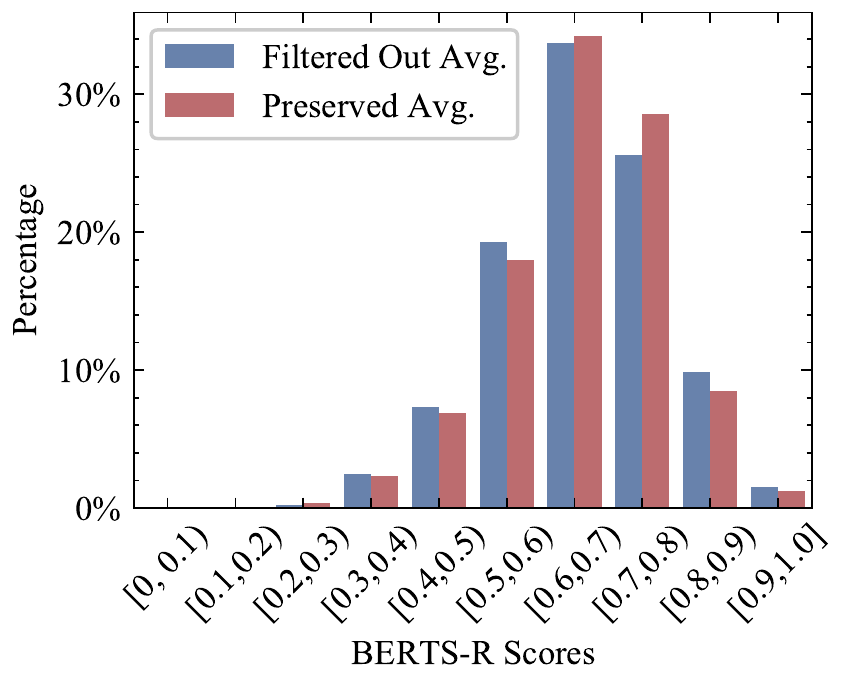} 
    \caption{Comparison of distribution of BERTS-R scores between filtered out and preserved sentences. Sentences with medium difficulty are more discriminative than which are either extremely hard or extremely easy.}
    \label{fig:score_distri}
    \vspace{-10pt}
\end{wrapfigure}
\paragraph{Translation Difficulty}
is not the same as the evaluation discernment as we assumed before. 
The test instances whose difﬁculty lies at the extremes of the scale are not useful in an evaluation aiming to distinguish between MT systems, so the preserved samples possibly are not the simplest or the most difficult instances.
Starting from this intuition, we investigate the distribution of the averaged score of the test instances on the WMT20 English$\rightarrow$German translation task since it involves competitive systems that are challenging for the automated evaluation \citep{freitag-etal-2020-bleu}. 
Obviously, Figure \ref{fig:score_distri} reveals that the preserved instances have moderate but not extreme difficulty.
The phenomenon that samples with slightly higher difficulty are preserved also conforms with the previous observation in terms of the sentence length, that the longer sentences are more vital in distinguishing the MT systems due to their strong capability, also echoes the previous research stating that this translation direction easily confused the automatic evaluation metric \citep{bojar-etal-2018-findings, barrault-etal-2019-findings}.

To conclude, a test item preserved by the proposed filtering method is discriminative in terms of its linguistic and data properties, thus the improvement in the correlation of the variance-aware test set is reasonable.
Moreover, the variance-aware filtering method has the potential for saving the human labor for diversifying the test sets.

\section{Conclusions and Future Work}
This paper introduces a method to select discriminative test instances from the machine translation benchmark and automatically create a series of variance-aware test sets.
Experimental results show that using the created test sets can improve the correlation performance of automatic evaluation results across representative test sets and languages, confirming the effectiveness and generality of the proposed method.
Further analysis of the features of the test instances supports the rationality of variance-aware test sets and ensures its reliability for other possible uses.

Future work includes: 1) investigating the use of the variance-aware test sets in other MT research questions. For example, using them as the validation sets in some time-consuming scenarios, e.g., neural architecture search and reinforcement learning; 2) applying the filtering method to the training set to accelerate the learning process; 3) extending the filtering method to other evaluation tasks like dialogue generation.

\section*{Acknowledgments}
This work was supported in part by the National Natural Science Foundation of China (Grant No. 61672555), the Science and Technology Development Fund, Macau SAR (Grant No. 0101/2019/A2), and the Multi-year Research Grant from the University of Macau (Grant No. MYRG2020-00054-FST). We thank the anonymous reviewers for their insightful comments.

\bibliographystyle{apalike}
\bibliography{neurips_data_2021}

\begin{thebibliography}{}

\bibitem[Bahdanau et~al., 2015]{DBLP:journals/corr/BahdanauCB14}
Bahdanau, D., Cho, K., and Bengio, Y. (2015).
\newblock Neural machine translation by jointly learning to align and
  translate.
\newblock In Bengio, Y. and LeCun, Y., editors, {\em 3rd International
  Conference on Learning Representations, {ICLR} 2015, San Diego, CA, USA, May
  7-9, 2015, Conference Track Proceedings}.

\bibitem[Banerjee and Lavie, 2005]{banerjee-lavie-2005-meteor}
Banerjee, S. and Lavie, A. (2005).
\newblock {METEOR}: An automatic metric for {MT} evaluation with improved
  correlation with human judgments.
\newblock In {\em Proceedings of the {ACL} Workshop on Intrinsic and Extrinsic
  Evaluation Measures for Machine Translation and/or Summarization}, pages
  65--72, Ann Arbor, Michigan. Association for Computational Linguistics.

\bibitem[Barrault et~al., 2019]{barrault-etal-2019-findings}
Barrault, L., Bojar, O., Costa-juss{\`a}, M.~R., Federmann, C., Fishel, M.,
  Graham, Y., Haddow, B., Huck, M., Koehn, P., Malmasi, S., Monz, C.,
  M{\"u}ller, M., Pal, S., Post, M., and Zampieri, M. (2019).
\newblock Findings of the 2019 conference on machine translation ({WMT}19).
\newblock In {\em Proceedings of the Fourth Conference on Machine Translation
  (Volume 2: Shared Task Papers, Day 1)}, pages 1--61, Florence, Italy.
  Association for Computational Linguistics.

\bibitem[Bawden et~al., 2020]{bawden-etal-2020-study}
Bawden, R., Zhang, B., Yankovskaya, L., T{\"a}ttar, A., and Post, M. (2020).
\newblock A study in improving {BLEU} reference coverage with diverse automatic
  paraphrasing.
\newblock In {\em Findings of the Association for Computational Linguistics:
  EMNLP 2020}, pages 918--932, Online. Association for Computational
  Linguistics.

\bibitem[Bojar et~al., 2018]{bojar-etal-2018-findings}
Bojar, O., Federmann, C., Fishel, M., Graham, Y., Haddow, B., Koehn, P., and
  Monz, C. (2018).
\newblock Findings of the 2018 conference on machine translation ({WMT}18).
\newblock In {\em Proceedings of the Third Conference on Machine Translation:
  Shared Task Papers}, pages 272--303, Belgium, Brussels. Association for
  Computational Linguistics.

\bibitem[Bojar et~al., 2017]{bojar-etal-2017-results}
Bojar, O., Graham, Y., and Kamran, A. (2017).
\newblock Results of the {WMT}17 metrics shared task.
\newblock In {\em Proceedings of the Second Conference on Machine Translation},
  pages 489--513, Copenhagen, Denmark. Association for Computational
  Linguistics.

\bibitem[Bojar et~al., 2016]{bojar-etal-2016-results}
Bojar, O., Graham, Y., Kamran, A., and Stanojevi{\'c}, M. (2016).
\newblock Results of the {WMT}16 metrics shared task.
\newblock In {\em Proceedings of the First Conference on Machine Translation:
  Volume 2, Shared Task Papers}, pages 199--231, Berlin, Germany. Association
  for Computational Linguistics.

\bibitem[Callison-Burch et~al., 2008]{callison2008further}
Callison-Burch, C., Fordyce, C.~S., Koehn, P., Monz, C., and Schroeder, J.
  (2008).
\newblock Further meta-evaluation of machine translation.
\newblock In {\em Proceedings of the third workshop on statistical machine
  translation}, pages 70--106.

\bibitem[Callison-Burch et~al., 2006]{callison-burch-etal-2006-evaluating}
Callison-Burch, C., Osborne, M., and Koehn, P. (2006).
\newblock Re-evaluating the role of {B}leu in machine translation research.
\newblock In {\em 11th Conference of the {E}uropean Chapter of the Association
  for Computational Linguistics}, Trento, Italy. Association for Computational
  Linguistics.

\bibitem[Chatterjee et~al., 2017]{chatterjee-etal-2017-guiding}
Chatterjee, R., Negri, M., Turchi, M., Federico, M., Specia, L., and Blain, F.
  (2017).
\newblock Guiding neural machine translation decoding with external knowledge.
\newblock In {\em Proceedings of the Second Conference on Machine Translation},
  pages 157--168, Copenhagen, Denmark. Association for Computational
  Linguistics.

\bibitem[Cho et~al., 2014]{cho-etal-2014-properties}
Cho, K., van Merri{\"e}nboer, B., Bahdanau, D., and Bengio, Y. (2014).
\newblock On the properties of neural machine translation: Encoder{--}decoder
  approaches.
\newblock In {\em Proceedings of {SSST}-8, Eighth Workshop on Syntax, Semantics
  and Structure in Statistical Translation}, pages 103--111, Doha, Qatar.
  Association for Computational Linguistics.

\bibitem[Denkowski and Lavie, 2014]{denkowski-lavie-2014-meteor}
Denkowski, M. and Lavie, A. (2014).
\newblock Meteor universal: Language specific translation evaluation for any
  target language.
\newblock In {\em Proceedings of the Ninth Workshop on Statistical Machine
  Translation}, pages 376--380, Baltimore, Maryland, USA. Association for
  Computational Linguistics.

\bibitem[Devlin et~al., 2019]{devlin-etal-2019-bert}
Devlin, J., Chang, M.-W., Lee, K., and Toutanova, K. (2019).
\newblock {BERT}: Pre-training of deep bidirectional transformers for language
  understanding.
\newblock In {\em Proceedings of the 2019 Conference of the North {A}merican
  Chapter of the Association for Computational Linguistics: Human Language
  Technologies, Volume 1 (Long and Short Papers)}, pages 4171--4186,
  Minneapolis, Minnesota. Association for Computational Linguistics.

\bibitem[Ding et~al., 2021a]{ding2021rejuvenation}
Ding, L., Wang, L., Liu, X., Wong, D.~F., Tao, D., and Tu, Z. (2021a).
\newblock Low-frequency words rejuvenation: Making the most of parallel data in
  non-autoregressive translation.
\newblock In {\em Proceedings of the 59th Annual Meeting of the Association for
  Computational Linguistics}, Online. Association for Computational
  Linguistics.

\bibitem[Ding et~al., 2021b]{ding2021understanding}
Ding, L., Wang, L., Liu, X., Wong, D.~F., Tao, D., and Tu, Z. (2021b).
\newblock Understanding and improving lexical choice in non-autoregressive
  translation.
\newblock In {\em International Conference on Learning Representations}.

\bibitem[Doddington, 2002]{doddington2002automatic}
Doddington, G. (2002).
\newblock Automatic evaluation of machine translation quality using n-gram
  co-occurrence statistics.
\newblock In {\em Proceedings of the second international conference on Human
  Language Technology Research}, pages 138--145.

\bibitem[Eriguchi et~al., 2019]{eriguchi2019incorporating}
Eriguchi, A., Hashimoto, K., and Tsuruoka, Y. (2019).
\newblock Incorporating source-side phrase structures into neural machine
  translation.
\newblock {\em Computational Linguistics}, 45(2):267--292.

\bibitem[Fadaee et~al., 2017]{fadaee-etal-2017-data}
Fadaee, M., Bisazza, A., and Monz, C. (2017).
\newblock Data augmentation for low-resource neural machine translation.
\newblock In {\em Proceedings of the 55th Annual Meeting of the Association for
  Computational Linguistics (Volume 2: Short Papers)}, pages 567--573,
  Vancouver, Canada. Association for Computational Linguistics.

\bibitem[Freitag et~al., 2020a]{freitag-etal-2020-human}
Freitag, M., Foster, G., Grangier, D., and Cherry, C. (2020a).
\newblock Human-paraphrased references improve neural machine translation.
\newblock In {\em Proceedings of the Fifth Conference on Machine Translation},
  pages 1183--1192, Online. Association for Computational Linguistics.

\bibitem[Freitag et~al., 2020b]{freitag-etal-2020-bleu}
Freitag, M., Grangier, D., and Caswell, I. (2020b).
\newblock {BLEU} might be guilty but references are not innocent.
\newblock In {\em Proceedings of the 2020 Conference on Empirical Methods in
  Natural Language Processing (EMNLP)}, pages 61--71, Online. Association for
  Computational Linguistics.

\bibitem[Guo and Hu, 2019]{guo-hu-2019-meteor}
Guo, Y. and Hu, J. (2019).
\newblock Meteor++ 2.0: Adopt syntactic level paraphrase knowledge into machine
  translation evaluation.
\newblock In {\em Proceedings of the Fourth Conference on Machine Translation
  (Volume 2: Shared Task Papers, Day 1)}, pages 501--506, Florence, Italy.
  Association for Computational Linguistics.

\bibitem[Hu et~al., 2019]{hu-etal-2019-domain-adaptation}
Hu, J., Xia, M., Neubig, G., and Carbonell, J. (2019).
\newblock Domain adaptation of neural machine translation by lexicon induction.
\newblock In {\em Proceedings of the 57th Annual Meeting of the Association for
  Computational Linguistics}, pages 2989--3001, Florence, Italy. Association
  for Computational Linguistics.

\bibitem[Kauchak and Barzilay, 2006]{kauchak-barzilay-2006-paraphrasing}
Kauchak, D. and Barzilay, R. (2006).
\newblock Paraphrasing for automatic evaluation.
\newblock In {\em Proceedings of the Human Language Technology Conference of
  the {NAACL}, Main Conference}, pages 455--462, New York City, USA.
  Association for Computational Linguistics.

\bibitem[Koehn and Knowles, 2017]{koehn-knowles-2017-six}
Koehn, P. and Knowles, R. (2017).
\newblock Six challenges for neural machine translation.
\newblock In {\em Proceedings of the First Workshop on Neural Machine
  Translation}, pages 28--39, Vancouver. Association for Computational
  Linguistics.

\bibitem[Lample and Conneau, 2019]{lample2019cross}
Lample, G. and Conneau, A. (2019).
\newblock Cross-lingual language model pretraining.
\newblock In {\em Advances in Neural Information Processing Systems},
  volume~32. Curran Associates, Inc.

\bibitem[L{\"a}ubli et~al., 2020]{laubli2020set}
L{\"a}ubli, S., Castilho, S., Neubig, G., Sennrich, R., Shen, Q., and Toral, A.
  (2020).
\newblock A set of recommendations for assessing human--machine parity in
  language translation.
\newblock {\em Journal of Artificial Intelligence Research}, 67:653--672.

\bibitem[Levenshtein, 1966]{levenshtein1966binary}
Levenshtein, V.~I. (1966).
\newblock Binary codes capable of correcting deletions, insertions, and
  reversals.
\newblock In {\em Soviet physics doklady}, volume~10, pages 707--710. Soviet
  Union.

\bibitem[Liu et~al., 2019]{liu-etal-2019-shared}
Liu, X., Wong, D.~F., Liu, Y., Chao, L.~S., Xiao, T., and Zhu, J. (2019).
\newblock Shared-private bilingual word embeddings for neural machine
  translation.
\newblock In {\em Proceedings of the 57th Annual Meeting of the Association for
  Computational Linguistics}, pages 3613--3622, Florence, Italy. Association
  for Computational Linguistics.

\bibitem[Lo, 2019]{lo-2019-yisi}
Lo, C.-k. (2019).
\newblock {Y}i{S}i - a unified semantic {MT} quality evaluation and estimation
  metric for languages with different levels of available resources.
\newblock In {\em Proceedings of the Fourth Conference on Machine Translation
  (Volume 2: Shared Task Papers, Day 1)}, pages 507--513, Florence, Italy.
  Association for Computational Linguistics.

\bibitem[Ma et~al., 2018]{ma-etal-2018-results}
Ma, Q., Bojar, O., and Graham, Y. (2018).
\newblock Results of the {WMT}18 metrics shared task: Both characters and
  embeddings achieve good performance.
\newblock In {\em Proceedings of the Third Conference on Machine Translation:
  Shared Task Papers}, pages 671--688, Belgium, Brussels. Association for
  Computational Linguistics.

\bibitem[Ma et~al., 2019]{ma-etal-2019-results}
Ma, Q., Wei, J., Bojar, O., and Graham, Y. (2019).
\newblock Results of the {WMT}19 metrics shared task: Segment-level and strong
  {MT} systems pose big challenges.
\newblock In {\em Proceedings of the Fourth Conference on Machine Translation
  (Volume 2: Shared Task Papers, Day 1)}, pages 62--90, Florence, Italy.
  Association for Computational Linguistics.

\bibitem[Mach{\'a}{\v{c}}ek and Bojar, 2013]{machacek-bojar-2013-results}
Mach{\'a}{\v{c}}ek, M. and Bojar, O. (2013).
\newblock Results of the {WMT}13 metrics shared task.
\newblock In {\em Proceedings of the Eighth Workshop on Statistical Machine
  Translation}, pages 45--51, Sofia, Bulgaria. Association for Computational
  Linguistics.

\bibitem[Mathur et~al., 2020]{mathur-etal-2020-results}
Mathur, N., Wei, J., Freitag, M., Ma, Q., and Bojar, O. (2020).
\newblock Results of the {WMT}20 metrics shared task.
\newblock In {\em Proceedings of the Fifth Conference on Machine Translation},
  pages 688--725, Online. Association for Computational Linguistics.

\bibitem[Nguyen and Chiang, 2018]{nguyen-chiang-2018-improving}
Nguyen, T. and Chiang, D. (2018).
\newblock Improving lexical choice in neural machine translation.
\newblock In {\em Proceedings of the 2018 Conference of the North {A}merican
  Chapter of the Association for Computational Linguistics: Human Language
  Technologies, Volume 1 (Long Papers)}, pages 334--343, New Orleans,
  Louisiana. Association for Computational Linguistics.

\bibitem[Papineni et~al., 2002]{papineni-etal-2002-bleu}
Papineni, K., Roukos, S., Ward, T., and Zhu, W.-J. (2002).
\newblock {B}leu: a method for automatic evaluation of machine translation.
\newblock In {\em Proceedings of the 40th Annual Meeting of the Association for
  Computational Linguistics}, pages 311--318, Philadelphia, Pennsylvania, USA.
  Association for Computational Linguistics.

\bibitem[Popovi{\'c}, 2012]{popovic-2012-morpheme}
Popovi{\'c}, M. (2012).
\newblock Morpheme- and {POS}-based {IBM}1 and language model scores for
  translation quality estimation.
\newblock In {\em Proceedings of the Seventh Workshop on Statistical Machine
  Translation}, pages 133--137, Montr{\'e}al, Canada. Association for
  Computational Linguistics.

\bibitem[Popovi{\'c}, 2015]{popovic-2015-chrf}
Popovi{\'c}, M. (2015).
\newblock chr{F}: character n-gram {F}-score for automatic {MT} evaluation.
\newblock In {\em Proceedings of the Tenth Workshop on Statistical Machine
  Translation}, pages 392--395, Lisbon, Portugal. Association for Computational
  Linguistics.

\bibitem[Popovic, 2019]{popovic-2019-reducing}
Popovic, M. (2019).
\newblock On reducing translation shifts in translations intended for {MT}
  evaluation.
\newblock In {\em Proceedings of Machine Translation Summit XVII Volume 2:
  Translator, Project and User Tracks}, pages 80--87, Dublin, Ireland. European
  Association for Machine Translation.

\bibitem[Rei et~al., 2020]{rei-etal-2020-comet}
Rei, R., Stewart, C., Farinha, A.~C., and Lavie, A. (2020).
\newblock {COMET}: A neural framework for {MT} evaluation.
\newblock In {\em Proceedings of the 2020 Conference on Empirical Methods in
  Natural Language Processing (EMNLP)}, pages 2685--2702, Online. Association
  for Computational Linguistics.

\bibitem[Sellam et~al., 2020]{sellam-etal-2020-bleurt}
Sellam, T., Das, D., and Parikh, A. (2020).
\newblock {BLEURT}: Learning robust metrics for text generation.
\newblock In {\em Proceedings of the 58th Annual Meeting of the Association for
  Computational Linguistics}, pages 7881--7892, Online. Association for
  Computational Linguistics.

\bibitem[Sennrich and Haddow, 2016]{sennrich-haddow-2016-linguistic}
Sennrich, R. and Haddow, B. (2016).
\newblock Linguistic input features improve neural machine translation.
\newblock In {\em Proceedings of the First Conference on Machine Translation:
  Volume 1, Research Papers}, pages 83--91, Berlin, Germany. Association for
  Computational Linguistics.

\bibitem[Sennrich et~al., 2016]{sennrich-etal-2016-neural}
Sennrich, R., Haddow, B., and Birch, A. (2016).
\newblock Neural machine translation of rare words with subword units.
\newblock In {\em Proceedings of the 54th Annual Meeting of the Association for
  Computational Linguistics (Volume 1: Long Papers)}, pages 1715--1725, Berlin,
  Germany. Association for Computational Linguistics.

\bibitem[Shen et~al., 2016]{shen-etal-2016-minimum}
Shen, S., Cheng, Y., He, Z., He, W., Wu, H., Sun, M., and Liu, Y. (2016).
\newblock Minimum risk training for neural machine translation.
\newblock In {\em Proceedings of the 54th Annual Meeting of the Association for
  Computational Linguistics (Volume 1: Long Papers)}, pages 1683--1692, Berlin,
  Germany. Association for Computational Linguistics.

\bibitem[Snover et~al., 2006]{Snover06astudy}
Snover, M., Dorr, B., Schwartz, R., Micciulla, L., and Makhoul, J. (2006).
\newblock A study of translation edit rate with targeted human annotation.
\newblock In {\em In Proceedings of Association for Machine Translation in the
  Americas}, pages 223--231.

\bibitem[Toral et~al., 2018]{toral-etal-2018-attaining}
Toral, A., Castilho, S., Hu, K., and Way, A. (2018).
\newblock Attaining the unattainable? reassessing claims of human parity in
  neural machine translation.
\newblock In {\em Proceedings of the Third Conference on Machine Translation:
  Research Papers}, pages 113--123, Brussels, Belgium. Association for
  Computational Linguistics.

\bibitem[Vaswani et~al., 2017]{DBLP:conf/nips/VaswaniSPUJGKP17}
Vaswani, A., Shazeer, N., Parmar, N., Uszkoreit, J., Jones, L., Gomez, A.~N.,
  Kaiser, L., and Polosukhin, I. (2017).
\newblock Attention is all you need.
\newblock In Guyon, I., von Luxburg, U., Bengio, S., Wallach, H.~M., Fergus,
  R., Vishwanathan, S. V.~N., and Garnett, R., editors, {\em Advances in Neural
  Information Processing Systems 30: Annual Conference on Neural Information
  Processing Systems 2017, December 4-9, 2017, Long Beach, CA, {USA}}, pages
  5998--6008.

\bibitem[Wieting et~al., 2019]{wieting-etal-2019-beyond}
Wieting, J., Berg-Kirkpatrick, T., Gimpel, K., and Neubig, G. (2019).
\newblock Beyond {BLEU}:training neural machine translation with semantic
  similarity.
\newblock In {\em Proceedings of the 57th Annual Meeting of the Association for
  Computational Linguistics}, pages 4344--4355, Florence, Italy. Association
  for Computational Linguistics.

\bibitem[Yankovskaya et~al., 2019]{yankovskaya-etal-2019-quality}
Yankovskaya, E., T{\"a}ttar, A., and Fishel, M. (2019).
\newblock Quality estimation and translation metrics via pre-trained word and
  sentence embeddings.
\newblock In {\em Proceedings of the Fourth Conference on Machine Translation
  (Volume 3: Shared Task Papers, Day 2)}, pages 101--105, Florence, Italy.
  Association for Computational Linguistics.

\bibitem[Zhan et~al., 2021]{zhan-etal-2021-difficulty}
Zhan, R., Liu, X., Wong, D.~F., and Chao, L.~S. (2021).
\newblock Difficulty-aware machine translation evaluation.
\newblock In {\em Proceedings of the 59th Annual Meeting of the Association for
  Computational Linguistics}, Online. Association for Computational
  Linguistics.

\bibitem[Zhang et~al., 2020]{zhang2019bertscore}
Zhang, T., Kishore, V., Wu, F., Weinberger, K.~Q., and Artzi, Y. (2020).
\newblock {BERTS}core: Evaluating text generation with {BERT}.
\newblock In {\em 8th International Conference on Learning Representations,
  {ICLR} 2020, Addis Ababa, Ethiopia, April 26-30, 2020}.

\end{thebibliography}

\newpage

\appendix

\section{Appendix}
\begin{wraptable}{r}{0.5\textwidth}
\vspace{-15pt}
\centering
\caption{Pearson correlations using original and variance-aware test sets on the competitive WMT20 Chinese $\leftrightarrow$ English tasks. Using variance-aware test sets generated by a random half of systems (+\textit{Half}) consistently improves the evaluation results on the remaining systems.}
\begin{tabular}{rccc}
\toprule
               & \textbf{~~~~Zh-En~~~~} & \textbf{~~~~En-Zh~~~~}  \\
\midrule
\textbf{BLEU}           & 0.957          & 0.888           \\
\textbf{\textit{+Half}} & \textbf{0.961} & \textbf{0.893}  \\
\hdashline
\textbf{COMET}          & 0.953          & -0.840          \\
\textbf{\textit{+Half}} & \textbf{0.963} & \textbf{-0.821} \\
\hdashline
\textbf{BLEURT}         & 0.941          & 0.803           \\
\textbf{\textit{+Half}} & \textbf{0.955} & \textbf{0.811}  \\
\hdashline
\textbf{BERTS-P}        & 0.938          & 0.902           \\
\textbf{\textit{+Half}} & \textbf{0.947} & \textbf{0.948}  \\
\hdashline
\textbf{BERTS-R}        & 0.956          & 0.934           \\
\textbf{\textit{+Half}} & \textbf{0.965} & \textbf{0.947}  \\
\hdashline
\textbf{BERTS-F}        & 0.949          & 0.924           \\
\textbf{\textit{+Half}} & \textbf{0.958} & \textbf{0.948} \\
\bottomrule
\end{tabular}
\label{table:robustness}
\vspace{-15pt}
\end{wraptable}

\paragraph{Robustness} A realistic scenario would be creating a filtered set with some existing systems and using the created test set to evaluate new systems. To verify the effectiveness of the proposed method in this scenario, we calculate the variance over a random half of the submitted systems of WMT20 Chinese$\leftrightarrow$English (Zh$\leftrightarrow$En) task and evaluate the generated test set on the remaining half. 
For example, the WMT20 Zh$\rightarrow$En task has 16 submitted results, we use a random half of the results (\textit{dong-nmt.1207, Tencent\_Translation.1249, WMTBiomedBaseline.183, Online-G.1569, DeepMind.381, Huawei\_TSC.889, Online-Z.1646, WeChat\_AI.1525}) for calculating variance and generating a variance-aware test set, and then evaluate the remained results (\textit{DiDiNLP.401, SJTU-NICT.320, Online-A.1585, zlabs-nlp.1176, Huoshan\_Translate.919, OPPO.1422, Online-B.1605, THUNLP.1498}) over the original test set and variance-aware test set. 
As shown in Table \ref{table:robustness}, the results indicate that the proposed method is still effective in such a scenario. 
Since the aim of this paper is to provide new test sets for the community to conduct evaluation for new systems, we calculate the variance of all submitted results to make the calculated variance more reliable and the generated test sets more discriminative.

\paragraph{Reproducibility} Since the organizers of WMT has released the generation results of the participant MT systems, we directly use these results to conduct the experiments instead of seeking for the original MT models. The data for reproducing the results including original test sets and human ratings, can be easily accessed from WMT official website as follows:
\begin{itemize}
	\item WMT16: \texttt{\href{http://www.statmt.org/wmt16/results.html}{http://www.statmt.org/wmt16/results.html}}. 
	\item WMT17: \texttt{\href{http://www.statmt.org/wmt17/results.html}{http://www.statmt.org/wmt17/results.html}}.
	\item WMT18: \texttt{\href{http://www.statmt.org/wmt18/results.html}{http://www.statmt.org/wmt18/results.html}}.
	\item WMT19: \texttt{\href{http://www.statmt.org/wmt19/results.html}{http://www.statmt.org/wmt19/results.html}}.
	\item WMT20: \texttt{\href{http://www.statmt.org/wmt20/results.html}{http://www.statmt.org/wmt20/results.html}}.
\end{itemize}

The evaluation metrics in this paper adopted the publicly available implementation:
\begin{itemize}
	\item BLEU: We used sacreBLEU implementation\footnote{https://github.com/mjpost/sacrebleu} and default hyperparameters. The evaluation signature is: \texttt{BLEU+case.mixed+lang+smooth.exp+tok.intl+version.1.4.14}. 
	\item COMET: We used original implementation\footnote{https://github.com/Unbabel/COMET}, \texttt{wmt-large-da-estimator-1719} evaluation model and default hyperparameters.
	\item BLEURT: We used original implementation\footnote{https://github.com/google-research/bleurt}, \texttt{BLEURT-base-128} checkpoint and default hyperparameters.
	\item BERTScore: used original implementation\footnote{https://github.com/Tiiiger/bert\_score}, default BERT model settings and hyperparameters.
\end{itemize}

\paragraph{Computational Resources} All the GPU computation is done by a single NVIDIA GeForce 1080Ti GPU card with CUDA Toolkit 10.1, and Intel Xeon E5-4655 CPU for handling other computation.

\paragraph{Data Licensing} The \textit{variance-aware test set} were created based on the original WMT test set. Thus, we follow the original data licensing plan already stated by WMT organizers, which is that ``The data released for the WMT news translation task can be freely used for research purposes, we just ask that you cite the WMT shared task overview paper, and respect any additional citation requirements on the individual data sets. For other uses of the data, you should consult with original owners of the data sets.'' (quoted from the ``LICENSING OF DATA'' part in the WMT official website\footnote{http://www.statmt.org/wmt20/translation-task.html}).

\paragraph{Accessibility and Maintenance} We have released the test sets and the codes for creating the variance-aware test sets on GitHub. Besides, we will maintain the repository including fixing any potential issues and updating the test sets if the new WMT test sets are released. Anyone can also prepare a specific variance-aware test set based on their customized data since the codes will be open-sourced.

\paragraph{Limitations} The major limitation of variance-aware filtering method is the evaluation quality of automatic metrics. If the metrics cannot give reasonable scores to the model's hypothesis, the calculated variance values could be meaningless. Although the BERTScore metric we used in this paper can evaluate the semantic-level overlap and is one of the state-of-the-art evaluation metrics, it is still inevitable to give inaccurate scores for the evaluation outside of general domain or document-level MT evaluation. In future works, it is reasonable to combine more kinds of evaluation scores to give a more accurate variance of test instances.

\end{document}